# A factorization criterion for acyclic directed mixed graphs

**Thomas S. Richardson**
Department of Statistics
University of Washington
tsr@stat.washington.edu

## Abstract

Acyclic directed mixed graphs, also known as semi-Markov models represent the conditional independence structure induced on an observed margin by a DAG model with latent variables. In this paper we present a factorization criterion for these models that is equivalent to the global Markov property given by (the natural extension of) d-separation.

## 1  Introduction

Acyclic directed mixed graphs, also known as semi-Markov models contain directed ($\rightarrow$) and bi-directed ($\leftrightarrow$) edges subject to the restriction that there are no directed cycles. The use of such graphs to represent a statistical model may be traced to path diagrams introduced by [14], where bi-directed edges are used to represent correlated errors. Such graphs are useful for representing the independence structure arising from a DAG model with hidden variables. More recently they have proved useful in characterizing the set of intervention distributions that are identifiable [4, 13]. Ancestral graphs [12] are a subclass of ADMGs.

The global Markov property for such models is given by the natural extension of d-separation to graphs with bi-directed edges. [11] provided a local Markov property for acyclic directed mixed graphs that was equivalent to the global property for all distributions. [5] introduced a weaker version of the local Markov property which is equivalent to the global property under additional assumptions.

In order to be able to score these models we require a parametrization of the set of distributions obeying the conditional independence relations given by an ADMG. Until now such parametrizations have only been available in the multivariate Gaussian case [2, 12].

In this paper we give a factorization result which leads directly to a parametrization in the multivariate binary case. It is not hard to extend the parameterization to the general discrete case, though we do not do it here for reasons of space.

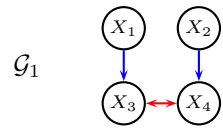

Figure 1: A simple ADMG $\mathcal{G}_1$. (Color is used to distinguish the two edge types.)

We illustrate the main ideas with the graph shown in Figure 1. This graph implies the conditional independence relations $X_1 \perp\!\!\!\perp X_2, X_4$ and $X_2 \perp\!\!\!\perp X_1, X_3$, which cannot be represented by a DAG model (without latents). The theorem given in this paper associates this graph with the following factorizations:

$$\begin{aligned}
p(x_1, x_2) &= p(x_1)p(x_2), \\
p(x_1, x_3) &= p(x_1)p(x_3 \mid x_1), \\
p(x_2, x_4) &= p(x_2)p(x_4 \mid x_2), \\
p(x_1, x_2, x_3) &= p(x_3 \mid x_1)p(x_1)p(x_2), \\
p(x_1, x_2, x_4) &= p(x_4 \mid x_2)p(x_1)p(x_2), \\
p(x_1, x_2, x_3, x_4) &= p(x_3, x_4 \mid x_1, x_2)p(x_1)p(x_2).
\end{aligned}$$

Note that the only terms here are: $p(x_1)$, $p(x_2)$, $p(x_3|x_1)$, $p(x_4|x_2)$ and $p(x_3, x_4|x_1, x_2)$. These factorizations lead to a (variation dependent) parametrization in the binary case: if the variables are binary then $p(X_1 = 0)$, $p(X_2 = 0)$, $p(X_3 = 0|x_1)$, $p(X_4 = 0|x_2)$ and $p(X_3 = X_4 = 0|x_1, x_2)$, with $x_1, x_2 \in \{0, 1\}$, are sufficient to specify any multivariate binary distribution obeying the global Markov property for $\mathcal{G}_1$; see [1]. Note that there are $1 + 1 + 2 + 2 + 4 = 10$ quantities here, which corresponds to the known dimension of this Markov model in the binary case.

It is possible to use the local directed Markov property for ADMGs in [11] (see §4.2), together with a given or-dering (consistent with the directed edges), to obtain factorizations in terms of univariate densities, as proposed in [9]. However, this strategy suffers from the following problem: if only one order is considered then the factorization corresponds to a DAG, and hence, in general, will not be sufficient capture all of the independencies given by the ADMG, e.g. ordering the vertices as they are numbered in $\mathcal{G}_1$ would give:

$$p(x_1, x_2, x_3, x_4) = p(x_1)p(x_2)p(x_3|x_1)p(x_4|x_1, x_2, x_3),$$

which does not imply $X_1 \perp\!\!\!\perp X_4$. On the other hand if all consistent orderings are considered then the set of densities does not form a parametrization as the dimension is too large, e.g. for $\mathcal{G}_1$ we would specify: $p(x_1)$, $p(x_2)$, $p(x_3 \mid x_1)$, $p(x_4 \mid x_2)$, $p(x_3 \mid x_1, x_2)$ and $p(x_4 \mid x_1, x_2)$, which gives 14 parameters in the binary case. Consequently these are not all free: some are constrained to be deterministic functions of others. The factorization in this paper is more complicated as it involves multivariate joint densities, but it has the advantage that it does give a parametrization.

In an ADMG, a set $A$ is said to be *ancestrally closed* if $x \to \cdots \to a$ in $\mathcal{G}$ with $a \in A$ implies that $x \in A$. Our main result is then as follows: a distribution $P$ obeys the global Markov property with respect to an ADMG $\mathcal{G}$ if and only if for every ancestrally closed set $A$:

$$p(X_A) = \prod_{H_i \in [A]_\mathcal{G}} p(X_{H_i} \mid X_{T_i})$$

where $[A]_\mathcal{G}$ denotes a partition of $A$ into sets $\{H_1, \ldots, H_k\}$, with corresponding conditioning sets $T_1, \ldots, T_k$; for details of the partition and conditioning sets see Theorem 4 below.

This generalizes and unifies three existing results for subclasses of ADMGs: First, the well-known factorization of an ancestrally closed set in a DAG $\mathcal{D}$ [10]:

$$p(X_A) = \prod_{v \in A} p(X_v \mid X_{\text{pa}_\mathcal{D}(v)}).$$

Second, the factorization associated with any set $A$ in a bi-directed graph $\mathcal{B}$ [3]:

$$p(X_A) = \prod_{\substack{C\,:\,C \subseteq A;\ C \text{ is connected in } \mathcal{B}, \\ \text{and } C \text{ is inclusion maximal in } A}} p(X_C);$$

where a set $C$ is *connected* if every pair of vertices in $C$ are joined by a path in $C$, and a connected set $C$ is *inclusion maximal* in $A$ if there is no set of $C^*$, with $C \subset C^* \subseteq A$, that is connected. (Note that in a bi-directed graph every set is ancestrally closed trivially since there are no directed edges.) Third the factorization associated with a bi-directed chain graph such as $\mathcal{G}_1$; see [1].

The paper is organized as follows: §2 contains basic definitions; §3 contains the factorization result; §4 outlines the proof.

## 2 Definitions and Basic Concepts

A *directed cycle* is a sequence of edges $x \to \cdots \to x$. If a directed mixed graph $\mathcal{G}$ contains no directed cycles then $\mathcal{G}$ is *acyclic*. There may be two edges between a pair of vertices in an acyclic directed mixed graph (ADMG), but in this case at least one edge must be bi-directed ($x \leftrightarrow y$): otherwise there would be a directed cycle (multiple edges of the same type are not permitted). The induced subgraph of $\mathcal{G}$ given by set $A$, denoted $\mathcal{G}_A$ consists of the subgraph of $\mathcal{G}$ with vertex set $A$, and those edges in $\mathcal{G}$ with endpoints in $A$. The *connected components* of a graph $\mathcal{G}$ are

For a vertex $x$ in a mixed graph $\mathcal{G}$, $\text{pa}_\mathcal{G}(x) \equiv \{v \mid v \to x \text{ in } \mathcal{G}\}$ is the set of *parents* of $x$; if $y \in \text{pa}_\mathcal{G}(x)$ then $x \in \text{ch}_\mathcal{G}(y)$, the set of *children* of $y$; $\text{sp}_\mathcal{G}(x) \equiv \{v \mid v \leftrightarrow x \text{ in } \mathcal{G}\}$ is the set of *spouses*[1] of $x$; $\text{an}_\mathcal{G}(x) \equiv \{v \mid v \to \cdots \to x \text{ in } \mathcal{G} \text{ or } v = x\}$ is the set of *ancestors* of $x$; similarly $\text{de}_\mathcal{G}(x) \equiv \{v \mid v \leftarrow \cdots \leftarrow x \text{ in } \mathcal{G} \text{ or } v = x\}$ is the set of *descendants* of $x$. These definitions are applied disjunctively to sets of vertices, so that, for example,

$$\text{pa}_\mathcal{G}(A) \equiv \bigcup_{x \in A} \text{pa}_\mathcal{G}(x), \qquad \text{sp}_\mathcal{G}(A) \equiv \bigcup_{x \in A} \text{sp}_\mathcal{G}(x).$$

(Note that $\text{sp}_\mathcal{G}(A) \cap A$ may be non-empty, and likewise for the other definitions.) A *path* between $x$ and $y$ in $\mathcal{G}$ is a sequence of edges $\langle \epsilon_1, \ldots, \epsilon_n \rangle$, such that there exists a sequence of distinct vertices $\langle x \equiv w_1, \ldots, w_{n+1} \equiv y \rangle$, $(n \geq 0)$, where edge $\epsilon_i$ has endpoints $w_i, w_{i+1}$ (paths consisting of a single vertex are permitted for the purpose of simplifying proofs). We denote a *subpath* of a path $\boldsymbol{\pi}$, by $\boldsymbol{\pi}(w_j, w_{k+1}) \equiv \langle \epsilon_j, \ldots, \epsilon_k \rangle$. It is necessary to specify a path as a sequence of edges rather than vertices because the latter does not specify a unique path when there may be two edges between a given pair of vertices. A path of the form $x \to \cdots \to y$ is a *directed path* from $x$ to $y$.

For a mixed graph $\mathcal{G}$ with vertex set $V$ we consider collections of random variables $(X_v)_{v \in V}$ taking values in probability spaces $(\mathfrak{X}_v)_{v \in V}$, where the probability spaces are either real finite-dimensional vector spaces or finite discrete sets. For $A \subseteq V$ we let $\mathfrak{X}_A \equiv \times_{v \in A} (\mathfrak{X}_v)$, $\mathfrak{X} \equiv \mathfrak{X}_V$ and $X_A \equiv (X_v)_{v \in A}$. We use the usual shorthand notation: $v$ denotes a vertex and a random variable $X_v$, likewise $A$ denotes a vertex set and $X_A$.

---

[1] Note that our usage, though established, differs from that of some authors who use the term 'spouse' to denote the other parents of the children of a vertex in a DAG.

## 2.1 The global Markov property for ADMGs

A non-endpoint vertex $z$ on a path is a *collider on the path* if the edges preceding and succeeding $z$ on the path have an arrowhead at $z$, i.e. $\to z \leftarrow$, $\leftrightarrow z \leftrightarrow$, $\leftrightarrow z \leftarrow$, $\to z \leftrightarrow$. A non-endpoint vertex $z$ on a path which is not a collider is a *non-collider on the path*, i.e. $\leftarrow z \to$, $\leftarrow z \leftarrow$, $\to z \to$, $\leftrightarrow z \to$, $\leftarrow z \leftrightarrow$. A path between vertices $x$ and $y$ in a mixed graph is said to be *m-connecting given a set $Z$* if

(i) every non-collider on the path is not in $Z$, and

(ii) every collider on the path is an ancestor of $Z$.

If there is no path m-connecting $x$ and $y$ given $Z$, then $x$ and $y$ are said to be *m-separated* given $Z$. Sets $X$ and $Y$ are said to be *m-separated* given $Z$, if for every pair $x$, $y$, with $x \in X$ and $y \in Y$, $x$ and $y$ are m-separated given $Z$.

A probability measure $P$ on $\mathfrak{X}$ is said to satisfy the *global Markov property* for $\mathcal{G}$ if for arbitrary disjoint sets $X, Y, Z$, ($Z$ may be empty) $X$ is m-separated from $Y$ given $Z$ in $\mathcal{G}$ implies $X \perp\!\!\!\perp Y \mid Z \, [P]$. Note that if $\mathcal{G}$ is a DAG then the above definition is identical to Pearl's d-separation criterion; see [6, 10].

In a given set $T$ with subset $A$, we will say that $B$ forms the *Markov blanket* for $A$ in $T$, if $B \subseteq T \setminus A$ and $A$ is m-separated from $T \setminus (B \cup A)$ given $B$, and there is no subset of $B$ for which this is true.

For an ADMG $\mathcal{G}$ we define:

$$\mathcal{A}(\mathcal{G}) \equiv \{A \mid \mathrm{an}_{\mathcal{G}}(A) = A\}$$

the set of *ancestrally closed sets*. The set $\mathcal{A}(\mathcal{G})$ forms a lattice so that if $A, B \in \mathcal{A}(\mathcal{G})$, then $A \cap B, A \cup B \in \mathcal{A}(\mathcal{G})$.

## 2.2 Barren subsets

For a given vertex set $A$ we define:

$$barren(A) \equiv \{x \mid x \in A; \mathrm{de}_{\mathcal{G}}(x) \cap A = \{x\}\}.$$

Note that this definitions is intrinsic to the set $A$ in the sense that a vertex $v \in \mathrm{barren}(A)$ if $v$ has no descendants in $\mathcal{G}$ that are *in $A$*; $v$ may have descendants in $\mathcal{G}$ that are not in $A$.

**Proposition 1.** *In an ADMG,* (i) $\mathrm{barren}(\mathrm{an}(A)) = \mathrm{barren}(A) \subseteq A$; (ii) *if* $A \in \mathcal{A}(\mathcal{G})$ *then* $\mathrm{an}(\mathrm{barren}(A)) = A$; (iii) *if* $B \subseteq A$ *then* $\mathrm{barren}(A) \cap B \subseteq \mathrm{barren}(B)$.

## 2.3 Districts

For a given ADMG $\mathcal{G}$, the *induced bi-directed graph* $(\mathcal{G})_{\leftrightarrow}$ is the graph formed by removing all directed edges from $\mathcal{G}$. Likewise $(\mathcal{G})_{\to}$ is the DAG formed by removing all bi-directed edges. The *district* (aka c-component) for a vertex $x$ in $\mathcal{G}$ is the connected component of $x$ in $(\mathcal{G})_{\leftrightarrow}$, or equivalently

$$dis_{\mathcal{G}}(x) \equiv \{y \mid y \leftrightarrow \cdots \leftrightarrow x \text{ in } \mathcal{G} \text{ or } x = y\}.$$

Similarly we let $\mathrm{dis}_A(x) = \mathrm{dis}_{\mathcal{G}_A}(x)$; if $x \notin A$ then we define $\mathrm{dis}_A(x) = \emptyset$. Note that $\mathrm{dis}_A(x) \subseteq \mathrm{dis}(x) \cap A$, with strict inclusion possible: a vertex $v$ is in $\mathrm{dis}_A(x)$ only if $v \in A$ and there is a path from $v$ to $x$ in $(\mathcal{G})_{\leftrightarrow}$ on which every vertex is in $A$; the condition for membership in $\mathrm{dis}(x) \cap A$ is the same, except that the requirement that every vertex on the path is in $A$ is removed. As usual we apply the definition disjunctively to sets:

$$\mathrm{dis}_A(B) = \bigcup_{x \in B} \mathrm{dis}_A(x).$$

**Proposition 2.** *If $W$ is a set in an ADMG $\mathcal{G}$, then:* (i) *the set of districts* $\{\mathrm{dis}_W(v) \mid v \in W\}$ *forms a partition of $W$;* (ii) *for any $v$,* $\mathrm{dis}_W(v) = \mathrm{dis}_{\mathrm{dis}_W(v)}(v)$; (iii) *vf $w \in W^* \subseteq W$ then* $\mathrm{dis}_{W^*}(w) \subseteq \mathrm{dis}_W(w)$.

A set $C$ is *path-connected* in $(\mathcal{G})_{\leftrightarrow}$ if every pair of vertices in $C$ are connected via a path in $(\mathcal{G})_{\leftrightarrow}$; equivalently, every vertex in $C$ has the same district in $\mathcal{G}$. Note that a set $C$ may be path-connected in $(\mathcal{G})_{\leftrightarrow}$ without forming a connected set in $(\mathcal{G})_{\leftrightarrow}$. (Recall that if a set $C$ is connected then between every pair of vertices in $C$ there is a path on which every vertex is itself in $C$; equivalently, every vertex in $C$ is in the same district in $\mathcal{G}_C$.)

A district $D = \mathrm{dis}_W(v)$ of a vertex $v$ in a set $W$ in $\mathcal{G}$ is said to be *ancestrally closed* if for every vertex $v$,

$$D = \mathrm{dis}_{\mathrm{an}_{\mathcal{G}}(D)}(v).$$

Thus the district of $v$ within $W$ would not get any larger if we were to add to $W$ all vertices in $\mathcal{G}$ that are ancestors of vertices that are already in the district of $v$ within $W$. (Note that if this property holds for one vertex $v \in D$, then it holds for all vertices in $D$.) By extension, a set $W$ will be said to have *ancestrally closed districts* if every district in $W$ is ancestrally closed.

**Proposition 3.** *If $W \in \mathcal{A}(\mathcal{G})$ then $W$ has ancestrally closed districts.*

Note however that the converse is false: a set may have ancestrally closed districts but not be ancestral. For example in the graph in Figure 2 all subsets of the vertices have ancestrally closed districts, including those which are not ancestral.

## 3 Factorization

The main result of the paper is as follows:

**Theorem 4.** *A probability distribution $P$ obeys the global Markov property for $\mathcal{G}$ if and only if for every $A \in \mathcal{A}(\mathcal{G})$,*

$$p(X_A) = \prod_{H \in [A]_\mathcal{G}} p(X_H \mid X_{\text{tail}(H)}) \tag{1}$$

where $[A]_\mathcal{G}$ denotes a partition of $A$ into sets $\{H_1, \ldots, H_k\} \subseteq \mathcal{H}(\mathcal{G})$, defined with $\text{tail}(H)$, below.

The factorization associated with an ADMG is of necessity more involved than in DAGs or chain graphs: in a DAG, one may always find a total ordering under which the parents of a vertex $v$ precede $v$. In chain graphs this is generalized so that the parents of an 'undirected component' precede the component itself. In the case of an ADMG graph, 'blocks' are *heads* (defined below) which are subsets of connected components of $\mathcal{G}_\leftrightarrow$. However, it is easy to find ADMG graphs in which there is no ordering under which every vertex that is a parent of at least one vertex in a given head, occurs prior to any vertex in that head. For example, in Figure 2, $\{1, 4\}$ and $\{2, 3\}$ form heads. It is easy to see that if 1 is ordered before 2 and 3, then we cannot also have 2 ordered before 1 and 4.

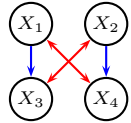

Figure 2: An ADMG in which there is no vertex ordering under which all parents of a head precede any vertex in the head.

The upshot of this is that the inclusion of a term in the factorization in which vertex $i$ is on the left of the conditioning bar, and $j$ is on the right, does not preclude the inclusion of a term, in the same factorization, in which the reverse holds. For example, we associate with the graph in Figure 2, the factorization:

$$p(x_1, x_2, x_3, x_4) = p(x_1, x_4 \mid x_2) p(x_2, x_3 \mid x_1). \tag{2}$$

It is easy to see that this factorization implies that $X_3 \perp\!\!\!\perp X_4 \mid X_1, X_2$. However, after marginalizing $x_3$ and $x_4$ it can be seen that (2) implies

$$p(x_1, x_2) = p(x_1 \mid x_2) p(x_2 \mid x_1),$$

so $X_1 \perp\!\!\!\perp X_2$, which also follows from the global Markov property associated with this graph. In this sense the factorization is 'cyclic' in nature.

### 3.1 Factorization terms: heads and tails

Each term in the factorization takes the form $p(X_H \mid X_T)$, $H, T \subseteq V$; $H \cap T = \emptyset$. Following [8], we will refer to $H$ as the *head* of the term $p(X_H \mid X_T)$, and $T$ as the *tail*. However, only certain pairs of sets $(H, T)$ will occur in factorizations associated with $\mathcal{G}$. An ordered pair of sets $(H, T)$ form the *head* and *tail* of a term associated with $\mathcal{G}$ if and only if all of the following hold:

(i) $H = \text{barren}(\text{an}_\mathcal{G}(H))$,

(ii) $H$ forms a path-connected set in $(\mathcal{G}_{\text{an}(H)})_\leftrightarrow$,

and

(iii) $T = (\text{dis}_{\text{an}(H)}(H) \setminus H) \cup \text{pa}(\text{dis}_{\text{an}(H)}(H))$.

Condition (i) may be re-expressed by saying that for every vertex $x \in H$, $\text{an}_\mathcal{G}(x) \cap H = \{x\}$. Condition (ii) is equivalent to the assertion that

$$\text{dis}_{\text{an}(H)}(H) = \bigcap_{v \in H} \text{dis}_{\text{an}(H)}(v).$$

For a graph $\mathcal{G}$, the set of *heads* is denoted by $\mathcal{H}(\mathcal{G})$. By condition (iii), if $H \in \mathcal{H}(\mathcal{G})$, then there is a unique tail which we denote $\text{tail}(H)$. The tail associated with $H$ consists of the variables in $\text{dis}_{\text{an}(H)}(H)$ that are not in $H$, and any vertex which is a parent of a vertex in $\text{dis}_{\text{an}(H)}(H)$, which includes all vertices in $\text{pa}(H)$. In what follows it will be useful to distinguish between these two subsets of $\text{tail}(H)$ hence we introduce the following definitions:

$$\begin{aligned} \textit{dis-tail}(H) &\equiv \text{dis}_{\text{an}(H)}(H) \setminus H, \\ \textit{pa-tail}(H) &\equiv \text{pa}(\text{dis}_{\text{an}(H)}(H)) \end{aligned}$$

(note that these two sets need not be disjoint). All vertices in the tail are connected to a vertex in $H$ by a path on which every non-endpoint vertex is a collider, and every vertex is an ancestor of some vertex in $H$. Given a set $W \supseteq \text{an}(H)$, with $\text{de}(W) \cap H = H$; $\text{sp}(\text{dis}_{\text{an}(H)}(H)) \cap W \subseteq \text{an}(H)$, the global Markov property for $\mathcal{G}$ implies that

$$H \perp\!\!\!\perp W \setminus (H \cup \text{tail}(H)) \mid \text{tail}(H)$$

or in other words, $\text{tail}(H)$ is the Markov blanket for $H$ in the set $W$.

For a bi-directed graph $\mathcal{B}$, $\mathcal{H}(\mathcal{B})$ is simply the set of connected subsets of vertices; for a DAG $\mathcal{D}$ it is the set of all singleton sets: $\mathcal{H}(\mathcal{D}) = \{\{v\} \mid v \in V\}$.

**Lemma 5.** *If $H \in \mathcal{H}(\mathcal{G})$, then $\text{tail}(H) \cap \text{de}(H) = \emptyset$.*

*Proof:* Suppose for a contradiction that there is some vertex $v \in \text{tail}(H) \cap \text{de}(H)$. Since, by definition,

$\mathrm{tail}(H) = \big(\mathrm{dis}_{\mathrm{an}(H)}(H) \setminus H\big) \cup \mathrm{pa}(\mathrm{dis}_{\mathrm{an}(H)}(H)) \subseteq \mathrm{an}(H)$, this implies that $H \neq \mathrm{barren}(\mathrm{an}_{\mathcal{G}}(H))$, which is a contradiction. □

**Corollary 6.** *If $H \in \mathcal{H}(\mathcal{G})$ then $H \cap \mathrm{tail}(H) = \emptyset$.*

*Proof:* Follows from Lemma 5 since $H \subseteq \mathrm{de}(H)$. □

In a complete ADMG $\mathcal{G}$, for every set $W \in \mathcal{A}(\mathcal{G})$, $\mathrm{barren}(W)$ is path-connected in $(\mathcal{G}_W)_{\leftrightarrow}$, so

$$\mathcal{H}(\mathcal{G}) = \{H \mid H = \mathrm{barren}(W) \text{ for some } W \in \mathcal{A}(\mathcal{G})\}.$$

**Proposition 7.** *If $A \in \mathcal{A}(\mathcal{G})$, then $\mathcal{H}(\mathcal{G}_A) = \{H \mid H \in \mathcal{H}(\mathcal{G}), H \subseteq A\}$.*

**Proposition 8.** *The mapping $\mathcal{H}(\mathcal{G}) \to \mathcal{A}(\mathcal{G})$ given by $A = \mathrm{an}_{\mathcal{G}}(H)$ is injective (one-to-one). Further, if $\mathcal{G}$ is complete then it is surjective (onto).*

### 3.2 Decomposition of a set: $[W]_{\mathcal{G}}$

For any ADMG $\mathcal{G}$ and a subset of vertices $W$, we define the following functions:

$$\Phi_{\mathcal{G}}(\emptyset) \equiv \emptyset,$$
$$\Phi_{\mathcal{G}}(W) \equiv$$
$$\left\{ H \,\Big|\, H = \bigcap_{x \in H} \mathrm{barren}\left(\mathrm{an}_{\mathcal{G}}\left(\mathrm{dis}_W(x)\right)\right); \; H \neq \emptyset \right\},$$

$$\psi_{\mathcal{G}}(W) \equiv W \setminus \bigcup_{H \in \Phi_{\mathcal{G}}(W)} H,$$

$$\psi_{\mathcal{G}}^{(k)}(W) \equiv \underbrace{\psi_{\mathcal{G}}(\cdots \psi_{\mathcal{G}}(W) \cdots)}_{k\text{-times}}, \qquad \psi_{\mathcal{G}}^{(0)}(W) \equiv W.$$

Finally, we define the *partition induced by $\mathcal{G}$ on $W$* to be:

$$[W]_{\mathcal{G}} \equiv \bigcup_{k \geq 0} \Phi_{\mathcal{G}}\left(\psi_{\mathcal{G}}^{(k)}(W)\right).$$

As described below, when applied to a set $W$, the function $\Phi_{\mathcal{G}}(W)$ returns a set of heads, one for each district in $W$. For a given district $D \subseteq W$, the associated head consists of those vertices in $D$ that are not ancestors (in $\mathcal{G}$) of any other vertices in $D$. The purpose of the intersection over elements in $H$ is to ensure that $H$ is path-connected in $(\mathcal{G}_W)_{\leftrightarrow}$; an equivalent definition is thus:

$$\Phi_{\mathcal{G}}(W) = \{H \mid \emptyset \neq H = \mathrm{barren}\left(\mathrm{an}_{\mathcal{G}}\left(\mathrm{dis}_W(H)\right)\right); \\ H \text{ is path-connected in } (\mathcal{G}_W)_{\leftrightarrow}\}.$$

Thus every pair of distinct vertices in $H$ are connected by a path in $(\mathcal{G}_W)_{\leftrightarrow}$ on which every edge is bi-directed, and is in $W$. As before it will not be true in general that $H$ itself will form a connected set in $(\mathcal{G}_W)_{\leftrightarrow}$.

The function $\psi_{\mathcal{G}}(W)$ simply removes from $W$ all those vertices which occur in some head in $\Phi_{\mathcal{G}}(W)$.

### 3.3 Parametrization

Multivariate binary distributions obeying the global Markov property with respect to an ADMG $\mathcal{G}$ may be parametrized by the following set of probabilities:

$$\left\{ p(X_H = 0 \mid X_{\mathrm{tail}(H)} = x_{\mathrm{tail}(H)}) \,\Big|\, H \in \mathcal{H}(\mathcal{G}), \right.$$
$$\left. x_{\mathrm{tail}(H)} \in \{0,1\}^{|\mathrm{tail}(H)|} \right\}.$$

Let $\alpha : V \mapsto \{0,1\}^{|V|}$ be an assignment of values to the variables indexed by $V$. Define $\alpha(W)$ to be the values assigned to variables indexed by a subset $W \subseteq V$. Let $\alpha^{-1}(0) = \{v \mid v \in V, \alpha(v) = 0\}$. Then:

$$p(X_V = \alpha(V)) = \sum_{C \,:\, \alpha^{-1}(0) \subseteq C \subseteq V} (-1)^{|C \setminus \alpha^{-1}(0)|} \times$$
$$\prod_{H \in [C]_{\mathcal{G}}} P\left(X_H = 0 \,\big|\, X_{\mathrm{tail}(H)} = \alpha(\mathrm{tail}(H))\right).$$

### 3.4 Examples

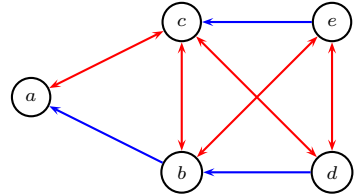

Figure 3: An ADMG used to illustrate recursive decompositions.

First consider the graph in Figure 2. We have the following decompositions:

$$\begin{aligned}
{[\{x_1, x_2\}]}_{\mathcal{G}} &= \{\{x_1\}, \{x_2\}\} \\
{[\{x_1, x_2, x_3\}]}_{\mathcal{G}} &= \{\{x_2, x_3\}, \{x_1\}\}, \\
{[\{x_1, x_2, x_4\}]}_{\mathcal{G}} &= \{\{x_1, x_4\}, \{x_2\}\}, \\
{[\{x_1, x_2, x_3, x_4\}]}_{\mathcal{G}} &= \{\{x_1, x_4\}, \{x_2, x_3\}\}, \\
{[\{x_1, x_3\}]}_{\mathcal{G}} &= \{\{x_1\}, \{x_3\}\}, \\
{[\{x_2, x_4\}]}_{\mathcal{G}} &= \{\{x_2\}, \{x_4\}\}.
\end{aligned}$$

Together with the singleton sets $\{x_1\}, \{x_2\}$ which have trivial decompositions, these represent all of the ancestral sets in this graph. These decompositions lead to the following factorizations:

$$\begin{aligned}
p(x_1, x_2) &= p(x_1)p(x_2), \\
p(x_1, x_2, x_3) &= p(x_2, x_3 | x_1)p(x_1), \\
p(x_1, x_2, x_4) &= p(x_1, x_4 | x_2)p(x_2), \qquad (3)\\
p(x_1, x_2, x_3, x_4) &= p(x_1, x_4 | x_2)p(x_2, x_3 | x_1), \\
p(x_1, x_3) &= p(x_3 | x_1)p(x_1), \\
p(x_2, x_4) &= p(x_4 | x_2)p(x_2).
\end{aligned}$$

Examples of the parametrization are as follows:

$$p(X_{1234}\!=\!0) = p(X_{14}\!=\!0|X_2\!=\!0)p(X_{23}\!=\!0|X_1\!=\!0),$$

where we use the obvious shorthand, e.g. $\{X_{14}\!=\!0\} \equiv \{X_1\!=\!X_4\!=\!0\}$;

$$\begin{aligned}&p(X_{12}\!=\!0, X_{34}\!=\!1)\\ &= p(X_1\!=\!0)p(X_2\!=\!0) - p(X_{23}\!=\!0|X_1\!=\!0)p(X_1\!=\!0)\\ &\quad - p(X_{14}\!=\!0|X_2\!=\!0)p(X_2\!=\!0)\\ &\quad + p(X_{14}\!=\!0|X_2\!=\!0)p(X_{23}\!=\!0|X_1\!=\!0);\end{aligned}$$

$$\begin{aligned}&p(X_{12}\!=\!1, X_{34}\!=\!0)\\ &= p(X_3\!=\!0|X_1\!=\!1)p(X_4\!=\!0|X_2\!=\!1)\\ &\quad - p(X_3\!=\!0|X_1\!=\!1)p(X_{14}\!=\!0|X_2\!=\!1)\\ &\quad - p(X_{23}\!=\!0|X_1\!=\!1)p(X_4\!=\!0|X_2\!=\!1)\\ &\quad + p(X_{23}\!=\!0|X_1\!=\!1)p(X_{14}\!=\!0|X_2\!=\!1).\end{aligned}$$

For a more complex example, we present five decompositions resulting from the graph in Figure 3:

$$\begin{aligned}{}[\{d,e\}]_{\mathcal{G}} &= \{\{d,e\}\},\\ [\{c,d,e\}]_{\mathcal{G}} &= \{\{c,d\},\{e\}\},\\ [\{b,c,d,e\}]_{\mathcal{G}} &= \{\{b,c\},\{d,e\}\},\\ [\{a,b,c,d,e\}]_{\mathcal{G}} &= \{\{a,c\},\{b,e\},\{d\}\},\\ [\{a,b,d,e\}]_{\mathcal{G}} &= \{\{a\},\{b,e\},\{d\}\},\end{aligned}$$

with the corresponding factorizations:

$$\begin{aligned}p(d,e) &= p(d,e),\\ p(c,d,e) &= p(c,d|e)p(e),\\ p(b,c,d,e) &= p(b,c|d,e)p(d,e),\\ p(a,b,c,d,e) &= p(a,c|b,d,e)p(b,e|d)p(d),\\ p(a,b,d,e) &= p(a|b)p(b,e|d)p(d).\end{aligned}$$

The first four do not imply any constraint on the associated margin. This is a consequence of the fact that there is only one district in each of these ancestral sets. Note that even though $\{a,b,c,d,e\}$ forms a single district, $\{a,b,d,e\}$ contains two districts.

### 3.5 Properties of $\Phi_{\mathcal{G}}(W)$

**Proposition 9.** *If $H \in \Phi_{\mathcal{G}}(W)$, then $H \subseteq W$.*

*Proof:* If for some $x \in H$, $\text{dis}_W(x) = \emptyset$ then $H \subseteq \text{barren}(\text{an}_{\mathcal{G}}(\emptyset)) = \emptyset$, which is a contradiction. But if $\text{dis}_W(x) \neq \emptyset$ then by definition $x \in W$. □

**Lemma 10.** *If $x_1, x_2 \in H \in \Phi_{\mathcal{G}}(W)$, then $\text{dis}_W(x_1) = \text{dis}_W(x_2)$, hence for all $x \in H$, $H = \text{barren}(\text{an}_{\mathcal{G}}(\text{dis}_W(x)))$.*

*Proof:* Let $D_i = \text{dis}_W(x_i)$, $i \in \{1,2\}$. By Proposition 9 $\text{dis}_W(x_i) \neq \emptyset$, $i \in \{1,2\}$. By Proposition 2 (i), if $D_1 \neq D_2$, then $D_1 \cap D_2 = \emptyset$. Further, by Proposition 1(i) $\text{barren}(\text{an}_{\mathcal{G}}(D_i)) \subseteq D_i$ $(i \in \{1,2\})$. Hence $D_1 \cap D_2 = \emptyset$ implies that $H = \emptyset$, which is a contradiction. □

**Corollary 11.** *If $H_1, H_2 \in \Phi_{\mathcal{G}}(W)$ and $H_1 \neq H_2$ then $\text{dis}_W(H_1) \cap \text{dis}_W(H_2) = \emptyset$.*

*Proof:* Suppose for a contradiction that $x \in \text{dis}_W(H_1) \cap \text{dis}_W(H_2)$. This implies that $\text{dis}_W(h_j) = \text{dis}_W(x)$, for any $h_j \in H_j$, $j \in \{1,2\}$. It then follows by Lemma 10 that

$$H_1 = \text{barren}(\text{an}_{\mathcal{G}}(\text{dis}_W(x))) = H_2,$$

which is a contradiction. □

**Corollary 12.** *If $H_1, H_2 \in \Phi_{\mathcal{G}}(W)$ and $H_1 \neq H_2$ then $H_1 \cap H_2 = \emptyset$.*

*Proof:* Immediate from Corollary 11, since $H_i \subseteq \text{dis}_W(H_i)$, $i \in \{1,2\}$. □

**Corollary 13.** *For any set $A$, $[A]_{\mathcal{G}}$ partitions $A$.*

*Proof:* By Corollary 12, no two sets in $\Phi_{\mathcal{G}}\left(\psi_{\mathcal{G}}^{(k)}(A)\right)$, for fixed $k$, overlap. Further, by Proposition 9, and the definition of $\psi_{\mathcal{G}}^{(k)}(A)$, the sets in $\Phi_{\mathcal{G}}\left(\psi_{\mathcal{G}}^{(i)}(A)\right)$ and $\Phi_{\mathcal{G}}\left(\psi_{\mathcal{G}}^{(j)}(A)\right)$ do not overlap. Since $W \neq \emptyset$ implies $\Phi_{\mathcal{G}}(W) \neq \emptyset$, $\{\psi_{\mathcal{G}}^{(i)}(A)\}_i$ is a monotonically decreasing sequence of sets, so the sets in $[A]_{\mathcal{G}}$ exhaust $A$. □

**Lemma 14.** *If $H \in \Phi_{\mathcal{G}}(W)$ then $H \in \mathcal{H}(\mathcal{G})$.*

*Proof:* We first show that $\text{barren}(\text{an}(H)) = H$. Since for any $h \in H$, $H = \text{barren}(\text{an}_{\mathcal{G}}(\text{dis}_W(h)))$, it follows by Proposition 1 (ii) that $\text{an}_{\mathcal{G}}(H) = \text{an}_{\mathcal{G}}(\text{dis}_W(h))$ hence $\text{barren}(\text{an}_{\mathcal{G}}(H)) = \text{barren}(\text{an}_{\mathcal{G}}(\text{dis}_W(h)))$. Since this holds for any $h \in H$, it then follows that

$$\text{barren}(\text{an}_{\mathcal{G}}(H)) = \bigcap_{h \in H} \text{barren}(\text{an}_{\mathcal{G}}(\text{dis}_W(h))) = H.$$

We now show that $\text{dis}_{\text{an}(H)}(H)$ is path-connected in $(\mathcal{G}_{\text{an}(H)})_{\leftrightarrow}$, or equivalently that

$$\text{dis}_{\text{an}_{\mathcal{G}}(H)}(H) = \bigcap_{v \in H} \text{dis}_{\text{an}_{\mathcal{G}}(H)}(v),$$

and thus $H \in \mathcal{H}(\mathcal{G})$. For any $h \in H$,

$$\begin{aligned}\text{dis}_{\text{an}_{\mathcal{G}}(H)}(h) &= \text{dis}_{\text{an}_{\mathcal{G}}(\text{dis}_W(h))}(h)\\ &\supseteq \text{dis}_{\text{dis}_W(h)}(h) = \text{dis}_W(h),\end{aligned}$$

where: the first equality is by Proposition 1(ii); the inclusion follows from Proposition 2 (iii); the last equality uses Proposition 2 (ii). Hence

$$\bigcap_{h \in H} \text{dis}_{\text{an}_{\mathcal{G}}(H)}(h) \supseteq \bigcap_{h \in H} \text{dis}_W(h) \supseteq H,$$

where the last inclusion follows from the definition of $\Phi_{\mathcal{G}}(W)$, and Proposition 1(i). By Proposition 2 (i) the LHS is either equal to the empty set or $\text{dis}_{\text{an}_{\mathcal{G}}(H)}(H)$, which concludes the proof. □

For $x \in A$ we define $head(x; A)$ to be the head $H$ containing $x$ in $[A]_{\mathcal{G}}$. For $H \in [A]_{\mathcal{G}}$ we refer to the $k \geq 0$, for which $H \in \Phi_{\mathcal{G}}(\psi^{(k)}(A))$, as the *depth* of $H$ in $A$, denoted $dep_A(H)$. Similarly the *depth of a vertex* $h \in A$, is the depth in $A$ of the head containing it: $dep_A(x) = dep_A(\text{head}(x; A))$.

### 3.6 Properties of $\psi_{\mathcal{G}}(W)$ when $W$ has ancestrally closed districts

We will show that if $W$ has ancestrally closed districts then so does $\psi_{\mathcal{G}}(W)$. We will also show that if $W$ has ancestrally closed districts then the district-tail of any head $H \in \Phi_{\mathcal{G}}(W)$ is contained in $\psi_{\mathcal{G}}(W)$.

In subsequent proofs we will make use of the following:

**Lemma 15.** *If $W$ has ancestrally closed districts and $H \in \Phi_{\mathcal{G}}(W)$ then $\text{dis}_W(v) = \text{dis}_{\text{an}(H)}(v)$, for all $v \in \text{dis}_W(H)$.*

*Proof:* First observe that $\text{an}(H) = \text{an}(\text{barren}(\text{an}(\text{dis}_W(H)))) = \text{an}(\text{dis}_W(H)) \supseteq \text{dis}_W(H)$. Hence $\text{dis}_{\text{an}(H)}(v)$ is not empty, if $v \in \text{dis}_W(H)$. Further,

$$\begin{aligned}\text{dis}_{\text{an}(H)}(v) &= \text{dis}_{\text{an}(\text{dis}_W(H))}(v) \\ &= \text{dis}_{\text{an}(\text{dis}_W(v))}(v) = \text{dis}_W(v),\end{aligned}$$

where the second equality follows since $v \in \text{dis}_W(H)$, the third equality follows since $W$ has ancestrally closed districts. □

**Lemma 16.** *If $W$ has ancestrally closed districts, then so does $\psi(W)$.*

**Corollary 17.** *If $A \in \mathcal{A}(\mathcal{G})$ then for $k \geq 0$, $\psi_{\mathcal{G}}^{(k)}(A)$ has ancestrally closed districts.*

*Proof:* The result follows from Proposition 3 and $k$-applications of Lemma 16. □

**Lemma 18.** *If $A \in \mathcal{A}(\mathcal{G})$, and $H \in [A]_{\mathcal{G}}$, with $dep_A(H) = k$, then $\text{dis-tail}(H) \subseteq \psi^{(k+1)}(A)$.*

*Proof:* By hypothesis $H \in \Phi_{\mathcal{G}}(\psi^{(k)}(A))$. By Corollary 17, $\psi^{(k)}(A)$ has ancestrally closed districts. Thus by Lemma 15, $\text{dis}_{\text{an}(H)}(H) = \text{dis}_{\text{an}(\psi^{(k)}(A))}(H)$. Further, since $\psi^{(k)}(A)$ has ancestrally closed districts, and $H \subseteq \psi^{(k)}(A)$, $\text{dis}_{\text{an}(\psi^{(k)}(A))}(H) = \text{dis}_{\psi^{(k)}(A)}(H)$. By Corollary 11, for $H' \in \Phi_{\mathcal{G}}(\psi^{(k)}(A))$, with $H' \neq H$, $\text{dis}_{\psi^{(k)}(A)}(H) \cap \text{dis}_{\psi^{(k)}(A)}(H') = \text{dis}_{\psi^{(k)}(A)}(H) \cap H' = \emptyset$.

Hence

$$\begin{aligned}\text{dis-tail}(H) &= \text{dis}_{\text{an}(H)}(H) \setminus H \\ &= \text{dis}_{\text{an}(H)}(H) \setminus \left(\bigcup_{H' \in \Phi_{\mathcal{G}}(\psi^{(k)}(A))} H'\right) \\ &= \text{dis}_{\psi^{(k)}(A)}(H) \setminus \left(\bigcup_{H' \in \Phi_{\mathcal{G}}(\psi^{(k)}(A))} H'\right) \\ &\subseteq \psi^{(k+1)}(A).\end{aligned}$$

Here the last inclusion follows by definition of $\psi(\cdot)$. □

## 4 Equivalence of global Markov property and factorization

An inductive argument establishing Theorem 4 faces several challenges: If $A$ is an ancestral set, there may be no head $H \in [A]_{\mathcal{G}}$ such that $A \setminus H$ is an ancestral set: see for example the set $\{x_1, x_2, x_3, x_4\}$ in Figure 2, in which $\{x_1, x_4\}$ and $\{x_2, x_3\}$ are the heads. Since Theorem 4 only makes assertions about factorizations over margins $p(X_A)$ where $A$ is an ancestral set, this means that we cannot apply an inductive argument where we add one term at a time. Further, as the examples in section 3.4 make clear, the heads occurring in the decomposition of an ancestral set $A$ do not necessarily occur in the decomposition of an ancestral superset $A \cup \{x\}$. Thus any inductive argument which hypothesizes that the factorization holds for $p(X_A)$ and then extends this to $p(X_{A \cup \{x\}})$ must also transform the terms in the factorization of $p(X_A)$.

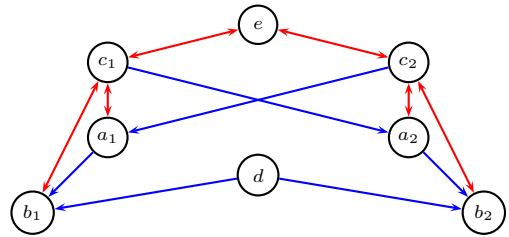

Figure 4: An ADMG used to illustrate the proof strategy for Theorem 4.

We first sketch our proof strategy by considering the example in Figure 4. Suppose, as an inductive hypothesis that, as required by Theorem 4,

$$\begin{aligned}&p(a_1, a_2, b_1, b_2, c_1, c_2, d) \\ &= p(b_1, c_1 | a_1, c_2, d) p(a_1 | c_2) p(b_2, c_2 | a_2, c_1, d) \\ &\quad \times p(a_2 | c_1) p(d).\end{aligned} \qquad (4)$$

We wish to show that the global Markov property implies that

$$p(a_1, a_2, b_1, b_2, c_1, c_2, d, e)$$
$$= p(e, b_1, b_2 | a_1, a_2, c_1, c_2, d)$$
$$\times p(d) p(a_1, c_1 | c_2) p(a_2, c_2 | c_1). \quad (5)$$

To this end we use the chain rule to further factorize the terms in equation (4), as follows:

$$p(b_1, c_1 | a_1, c_2, d) = p(b_1 | a_1, c_1, c_2, d) p(c_1 | a_1, c_2, d),$$
$$p(b_2, c_2 | a_2, c_1, d) = p(b_2 | a_2, c_1, c_2, d) p(c_2 | a_2, c_1, d).$$

Since $c_1$ is m-separated from $d$ given $a_1$ and $c_2$, $p(c_1 | a_1, c_2, d) = p(c_1 | a_1, c_2)$. Likewise $p(c_2 | a_2, c_1, d) = p(c_2 | a_2, c_1)$. Also since $b_1$ is m-separated from $a_2$ given $\{a_1, c_1, c_2, d\}$ we have $p(b_1 | a_1, c_1, c_2, d) = p(b_1 | a_1, a_2, c_1, c_2, d)$. Finally, we note that $b_2$ is m-separated from $\{a_1, b_1\}$ given $\{a_2, c_1, c_2, d\}$, hence $p(b_2 | a_2, c_1, c_2, d) = p(b_2 | a_1, a_2, b_1, c_1, c_2, d)$. We now obtain:

$$p(e | a_1, a_2, b_1, b_2, c_1, c_2, d) p(a_1, a_2, b_1, b_2, c_1, c_2, d)$$
$$= \Big( p(e | a_1, a_2, b_1, b_2, c_1, c_2, d)$$
$$\times p(b_1 | a_1, a_2, c_1, c_2, d) p(b_2 | a_1, a_2, b_1, c_1, c_2, d) \Big)$$
$$\times p(d) \Big( p(c_1 | a_1, c_2) p(a_1 | c_2) \Big)$$
$$\times \Big( p(c_2 | a_2, c_1) p(a_2 | c_1) \Big)$$

which establishes the required factorization (5). A key step in this argument was the ordering used to decompose $p(b_1, c_1 | a_1, c_2, d)$ into univariate marginals. Had this been decomposed into $p(c_1 | a_1, b_1, c_2, d) p(b_1 | a_1, c_2, d)$, the argument would not have gone through, as the global Markov property could not have been applied, and the terms could not have been re-arranged in the required manner.

We first consider the relation between the depth of a vertex in $A$, and in $A \cup \{x\}$, (in the case where both sets are ancestral). We show that there exists a (not necessarily unique) total order on the vertices in $A \cup \{x\}$, which is compatible with both orderings on the heads, in the sense that $a$ precedes $b$ if the head containing $a$ precedes the head containing $b$ in $[A]_\mathcal{G}$, or in $[A \cup \{x\}]_\mathcal{G}$. We use this total order to factor the head terms in the factorization of $p(x_A)$, given by the inductive hypothesis, into univariate terms. We will then apply the global Markov property to add or remove terms to the conditioning sets as required. Finally, regrouping terms will yield the required factorization of $p(x_{A \cup \{x\}})$.

**Lemma 19.** *If $A, A \cup \{x\} \in \mathcal{A}(\mathcal{G})$, $(x \notin A)$ then for $w \in A$, $\text{dep}_A(w) \leq \text{dep}_{A \cup \{x\}}(w)$.*

*Proof:* By induction on $k = \text{dep}_{A \cup \{x\}}(w)$. □

**Lemma 20.** *If $A, A \cup \{x\} \in \mathcal{A}(\mathcal{G})$, $(x \notin A)$ then for $w \in A$, $\text{dep}_{A \cup \{x\}}(w) \leq \text{dep}_A(w) + 1$.*

A total ordering $<$ on the vertices in a set $A \in \mathcal{A}(\mathcal{G})$ will be said to be *consistent with the depth ordering induced by $A$*, if whenever $\text{dep}_A(b) < \text{dep}_A(c)$, then $b < c$.

**Corollary 21.** *If $A, A \cup \{x\} \in \mathcal{A}(\mathcal{G})$, $(x \notin A)$ then there exists a total ordering on the vertices in $A$ that is consistent with the depth ordering induced by $A$ **and** the depth ordering induced by $A \cup \{x\}$.*

We note in passing that this Corollary cannot, in general, be extended to ancestral sets differing by more than one vertex. Consider the sets $A = \{c, d, e\}$, and $A^* = \{a, b, c, d, e\}$ in Figure 3. $\text{dep}_A(e) = 1 > \text{dep}_A(d) = 0$, but $\text{dep}_{A^*}(d) = 2 > \text{dep}_{A^*}(e) = 1$.

If $<$ is a total ordering, define $suc_<(x) = \{v \mid v \geq x\}$.

### 4.1 Markov blankets derived from a depth ordering

**Lemma 22.** *Let $H \in [A]_\mathcal{G}$ and $H = \{h_{i_1}, \ldots, h_{i_k}\}$ where the vertices are ordered according to a depth consistent total ordering $<$. Let $T_j = \{h_{i_j}, \ldots, h_{i_k}\} \cup \text{tail}(H)$. Then*

$$B_j = \Big( \text{dis}_{suc_<(h_{i_j})}(h_{i_j}) \setminus \{h_{i_j}\} \Big) \cup \text{pa}(\text{dis}_{suc_<(h_{i_j})}(h_{i_j}))$$

*forms the Markov blanket of $\{h_{i_j}\}$ in $T_j$.*

Note that $B_j$ is solely determined by $h_{i_j}$, the ordering, and the graph $\mathcal{G}$. Thus, if $A$ and $A \cup \{x\}$ are ancestral sets $(x \notin A)$ and for some vertex $v \in A$, $\text{head}(v; A) = H$, $\text{head}(v; A \cup \{x\}) = H^*$, with associated tails $T$ and $T^*$, then, under the ordering given in Corollary 21, the Markov blankets of $\{v\}$ in the sets $(H \cup T) \cap suc_<(h)$ and $(H^* \cup T^*) \cap suc_<(h)$ are the same.

### 4.2 The ordered local Markov property for ADMGs

A total ordering ($\prec$) on the vertices of $\mathcal{G}$, which satisfies $x \prec y \Rightarrow y \notin \text{an}(x)$ is said to be *consistent with the ancestor relations in $\mathcal{G}$*. Let $\text{pre}_{\mathcal{G}, \prec}(x) \equiv \{v \mid v \prec x \text{ or } v = x\}$. If $A \in \mathcal{A}(\mathcal{G})$ and $x \in \text{barren}(A)$ then the Markov blanket of $\{x\}$ with respect to the induced subgraph on $A$ is

$$\text{mb}(x, A) \equiv \text{pa}_\mathcal{G}\Big(\text{dis}_A(x)\Big) \cup \Big(\text{dis}_A(x) \setminus \{x\}\Big).$$

A probability measure $P$ satisfies the *ordered local Markov property* for $\mathcal{G}$, with respect to the ancestor consistent ordering $\prec$, if for any $x$, and ancestral set $A$ such that $x \in A \subseteq \text{pre}_{\mathcal{G},\prec}(x)$,

$$\{x\} \perp\!\!\!\perp A \setminus (\text{mb}(x,A) \cup \{x\}) \mid \text{mb}(x,A) \quad [P].$$

Note that $\text{ch}_{\mathcal{G}}(x) \cap \text{pre}_{\mathcal{G},\prec}(x) = \emptyset$, so $\text{ch}_{\mathcal{G}}(x) \cap A = \emptyset$. Since in a DAG, for an arbitrary ancestral set $A$ containing no children of $x$, $\text{mb}(x,A) = \text{pa}(x)$, the ordered local Markov property given above reduces to the local well-numbering Markov property introduced by [7]. In [11] it is shown that a probability measure obeys the global Markov property for an ADMG $\mathcal{G}$ if and only if it obeys the ordered local Markov property for $\mathcal{G}$.

**Lemma 23.** *If $P$ factorizes according to $\mathcal{G}$, then $P$ obeys the ordered local Markov property for $\mathcal{G}$.*

*Proof:* Let $x \in A \in \mathcal{A}(\mathcal{G})$, with $A \subseteq \text{pre}_{\mathcal{G},\prec}(x)$. Let $H = \text{head}(x;A)$. It is sufficient to note that $\text{mb}(x;A) = \text{tail}(H) \cup (H \setminus \{x\})$. $\square$

### 4.3 Proof of Theorem 4

*Proof:* It follows directly from Lemma 23, and the results in [11] that if a distribution factorizes according to (1) then it obeys the global Markov property for $\mathcal{G}$.

To prove that if $\mathcal{G}$ obeys the global Markov property then it factorizes according to (1) we proceed by induction on the size of the ancestral set $A \in \mathcal{A}(\mathcal{G})$.

If $|A| = 1$, then the claim is clearly trivial.

If $|A| > 1$, then $\text{barren}(A) \neq \emptyset$. Let $w$ be a vertex in $\text{barren}(A)$, hence $A \setminus \{w\} \in \mathcal{A}(\mathcal{G})$. Let $A^\dagger = A \setminus \{w\}$. By the induction hypothesis, (1) holds for $p(x_{A^\dagger})$. By Corollary 21 there exists a total ordering ($<$) on the vertices in $A^\dagger$ that is consistent with the depth orderings induced by $A^\dagger$ and $A$. Let $v$ be an arbitrary vertex in $A^\dagger$. Let $H = \text{head}(v;A)$ and $H^\dagger = \text{head}(v;A^\dagger)$. Further let $H = \{h_1, \ldots h_k\}$, and $H^\dagger = \{h_1^\dagger, \ldots h_{k^\dagger}^\dagger\}$, where the vertices are sequenced according to $<$. (The vertices in $H$ and $H^\dagger$ will, in general, be subsequences of the sequence of vertices totally ordered by $<$. However, we refrain from using double subscripts in order to keep the notation simple.) Hence $v = h_j = h_{j^\dagger}^\dagger$, for some $j, j^\dagger$. By Lemma 22, the Markov blankets of $\{v\}$ in the sets $T_j = \{h_j, \ldots, h_k\} \cup \text{tail}(H)$ and $T_j^\dagger = \{h_j^\dagger, \ldots, h_{k^\dagger}^\dagger\} \cup \text{tail}(H^\dagger)$ are the same. Hence

$$p(v \mid h_{j+1}, \ldots, h_k, \text{tail}(H))$$
$$= p\left(v \mid h_{j^\dagger+1}^\dagger, \ldots, h_{k^\dagger}^\dagger, \text{tail}(H^\dagger)\right)$$

(where the $h$'s are omitted if $v = h_k$ and the $h^\dagger$'s are omitted if $v = h_k^\dagger$). Thus the density over the $A^\dagger$ margin may be factorized into univariate terms of the appropriate form. Finally, we observe that since $w \in \text{barren}(A)$, $w$ is m-separated from $A \setminus (\text{head}(w;A) \cup \text{tail}(w;A))$ given $(\text{head}(w;A) \cup \text{tail}(w;A)) \setminus \{w\}$, so

$$p(w \mid A \setminus \{w\})$$
$$= p(w \mid (\text{head}(w;A) \cup \text{tail}(w;A)) \setminus \{w\}).$$

This completes the proof. $\square$

### Acknowledgements

This research was supported by U.S. NSF grant DMS-0505865 and NIH grant R01 AI032475.